\documentclass[10pt,twocolumn,letterpaper]{article}
\usepackage{setspace}
\usepackage{cvpr}              

\newcommand{\TODO}[1]{\textbf{\color{red}[TODO: #1]}}
\usepackage{pgfplots}
\pgfplotsset{compat=1.18}
\usepackage{subcaption} 
\usepackage{booktabs}
\usepackage{multirow}
\usepackage[accsupp]{axessibility} 

\renewcommand{\TODO}[1]{}

\usepackage{microtype}

\renewcommand{\paragraph}[1]{\vspace{.5em}\noindent\textbf{#1.}}

\definecolor{cvprblue}{rgb}{0.21,0.49,0.74}
\usepackage[pagebackref,breaklinks,colorlinks,allcolors=cvprblue]{hyperref}

\def\paperID{15} 
\def\confName{CVPR}
\def\confYear{2026}

\title{Beyond the Fold: Quantifying Split-Level Noise and the Case for Leave-One-Dataset-Out AU Evaluation}

\author{Saurabh Hinduja\\
CGI Technologies and Solutions Inc\\
{\tt\small saurabhh@ntweat.com}
\and
Gurmeet Kaur\\
CGI Technologies and Solutions Inc\\
{\tt\small gurmeet.usf@gmail.com}
\and
Maneesh Bilalpur\\
University of Pittsburgh\\
{\tt\small mab623@pitt.edu}
\and
Jeffrey F. Cohn\\
University of Pittsburgh\\
{\tt\small jeffcohn@pitt.edu}
\and
Shaun Canavan\\
University of South Florida\\
{\tt\small scanavan@usf.edu}
}

\begin{document}
\setstretch{0.93}
\maketitle
\begin{abstract}
Subject-exclusive cross-validation is the standard evaluation protocol for facial Action Unit (AU) detection, yet reported improvements are often small. We show that cross-validation itself introduces measurable stochastic variance. On BP4D+, repeated 3-fold subject-exclusive splits produce an empirical noise floor of $\pm 0.065$ in average F1, with substantially larger variation for low-prevalence AUs. Operating-point metrics such as F1 fluctuate more than threshold-independent measures such as AUC, and model ranking can change under different fold assignments.

We further evaluate cross-dataset robustness using a Leave-One-Dataset-Out (LODO) protocol across five AU datasets. LODO removes partition randomness and exposes domain-level instability that is not visible under single-dataset cross-validation. Together, these results suggest that 
gains often reported in cross-fold validation may fall within protocol variance.  Leave-one-dataset-out cross-validation yields more stable and interpretable findings.
\end{abstract}

\section{Introduction}
\label{sec:introduction}
Automated Facial Action Unit (AU) detection is a central task in affective computing and behavioral analysis. Rooted in the Facial Action Coding System (FACS) introduced by Ekman and Friesen \citep{Ekman2002FACS}, AU detection decomposes facial behavior into anatomically interpretable muscle activations. Since the early work of Tian \emph{et al.} \citep{Tian2001RecognizingAUs}, the field has evolved from handcrafted descriptors to convolutional architectures, graph-based models, and transformer-based approaches evaluated on benchmarks such as BP4D+ \citep{zhang2016multimodal}. Despite rapid architectural progress, the dominant evaluation paradigm has remained largely unchanged: subject-exclusive $k$-fold cross-validation on a single dataset. Under this protocol, models are trained on a subset of subjects and evaluated on held-out subjects drawn from the same dataset distribution. Reported improvements are frequently modest, often on the order of $+0.01$ to $+0.02$ in average F1 score \cite{hinduja2024time}, yet such deltas are routinely interpreted as evidence of state-of-the-art (SOTA) advancement. Implicit in this practice is the assumption that cross-validation provides a stable and reliable estimate of model performance.

In this work, we challenge that assumption. We demonstrate that subject-exclusive cross-validation itself introduces measurable stochastic variance in AU detection performance. Even when the dataset and architecture are held fixed, altering only the assignment of subjects to folds can yield substantial changes in reported metrics. We refer to this phenomenon as split-level noise. Throughout this paper, we use the term \emph{noise} to mean protocol-induced variability in reported performance that arises solely from random subject partitioning. Because current AU datasets possess limited subject counts, they lack the large sample sizes required to yield mathematically stable performance estimates, which is a fundamental limitation in machine learning evaluation \cite{SAIDI2025101185}. To formally quantify this uncertainty, we treat cross-validation performance as a random variable and establish an empirical noise floor corresponding to a 95\% confidence interval ($1.96\sigma$). On BP4D+, repeated cross-validation reveals a 95\% noise margin of $\pm0.065$ F1, a magnitude that frequently exceeds the incremental gains reported in recent literature.

Our contribution differs from prior critiques of AU evaluation metrics. Jeni \emph{et al.} \citep{jeni2013_facing_imbalanced} showed that skewed (imbalanced) data can bias common performance measures, emphasizing careful metric selection. More recently, Hinduja \emph{et al.} \citep{hinduja2024time} argued that F1-binary should be retired for AU detection because it is highly sensitive to class prevalence. These works establish that operating-point metrics can follow \emph{dataset distribution}. In contrast, we identify and quantify a distinct instability source: even when the dataset distribution is unchanged, evaluation outcomes can follow \emph{subject partitioning}. That is, protocol-induced randomness in fold composition shifts AU base rates and co-occurrence statistics within the test fold, producing performance variation unrelated to genuine methodological improvement.

Split-level noise also interacts with broader cross-domain generalization concerns. Ertugrul \emph{et al.} \citep{crossing_domains_au_coding_pmc} highlighted the difficulty of transferring AU models across datasets, demonstrating that performance degrades under distribution shift. Our findings suggest that instability arises at two complementary levels: (i) \emph{partition-level} stochasticity within a dataset, and (ii) \emph{dataset-level} distribution shift across domains. The former has received little quantitative attention. As a result, many published comparisons rely on single cross-validation runs without variance reporting, potentially conflating fold-specific artifacts with genuine architectural gains. Metric choice also matters: operating-point metrics such as F1 exhibit greater partition sensitivity than threshold-independent measures such as AUC, motivating the reporting of both.

To address these issues, we propose Leave-One-Dataset-Out (LODO) evaluation as a principled alternative. Under LODO, models are trained on multiple heterogeneous datasets and evaluated on a completely unseen dataset. This protocol eliminates intra-dataset split lotteries and reframes progress in terms of domain-level generalization. Rather than measuring how well a model fits a particular fold configuration, LODO evaluates its ability to reconcile multiple domains and generalize to an unseen distribution.

\paragraph{Contributions.}
\begin{itemize}
    \item The statistical uncertainty of AU evaluation caused by limited dataset sample sizes is formalized. By treating cross-validation performance as a random variable, we establish an empirical noise floor based on 95\% confidence intervals ($1.96\sigma$), demonstrating F1 instability under repeated splits on BP4D+ that frequently obscures reported state-of-the-art gains.
    \item We show that operating-point metrics (e.g., F1) show greater partition sensitivity than AUC, and
    that model ranking can change under different fold constructions.
    \item Leave-One-Dataset-Out (LODO) evaluation is introduced along with subject-level bootstrapping to isolate true domain shifts from sampling variability. This establishes a statistically grounded multi-dataset baseline as an alternative to intra-dataset cross-validation.
\end{itemize}

\section{Related Work}

\subsection{Modern Architectures for AU Detection}

Recent progress in facial Action Unit (AU) detection has been driven by structured deep architectures that explicitly model inter-AU relationships. Graph neural networks have emerged as a dominant paradigm. Luo \emph{et al.} proposed Multi-Edge Graph AU (ME-GraphAU)~\cite{Luo2022MEGraphAU}, which learns multiple types of dependencies between AUs via message passing. Song \emph{et al.} introduced Hybrid Message Passing (HMP)~\cite{Song2021HMP}, enabling adaptive relational reasoning among AUs. More recently, Wang \emph{et al.} presented Multi-Scale Dynamic and Hierarchical Relationship Modeling (MDHR)~\cite{Wang2024MDHR}, incorporating hierarchical and temporal dependencies to better reflect anatomical structure.

Transformer-based approaches have further advanced intra-dataset performance. Yuan \emph{et al.}~\cite{auformer_2024} demonstrated parameter-efficient AU detection using Vision Transformers. MAE-Face~\cite{Zhao2023MAEFace} applies masked autoencoder pretraining to face representation learning and improves downstream AU recognition under limited labeled data. VisAULa~\cite{visaula_2026} incorporates token-level alignment within transformer architectures to refine fine-grained AU localization. Despite architectural sophistication, these models are almost universally evaluated using subject-exclusive cross-validation within a single dataset, typically reporting mean performance without repeated partition analysis.

\subsection{Self-Supervised and Foundation-Models}

Limited AU annotations have motivated self-supervised learning (SSL) and foundation-model strategies. Chang and Wang proposed knowledge-driven self-supervised representation learning (KDSSL)~\cite{Chang2022KDSSL}, leveraging FACS priors to guide representation learning. AU-vMAE~\cite{auvmae_2024} extends masked autoencoding into the video domain for temporal AU modeling. More recently, multimodal approaches have begun integrating textual supervision. Li \emph{et al.} introduced hierarchical vision-language interaction mechanisms to align textual AU descriptions with visual representations~\cite{hierarchical_vl_interaction_2026}, reflecting a broader trend toward vision–language pretraining and large-model adaptation  facial behavior analysis. While these approaches improve representation capacity, their evaluation protocols remain predominantly intra-dataset and partition-fixed.

\subsection{Metrics and Sensitivity to Distribution}

Evaluation robustness has been examined primarily from a metric perspective. For example, Jeni \emph{et al.}~\cite{jeni2013_facing_imbalanced} showed that skewed AU prevalence can bias commonly used metrics. Hinduja \emph{et al.}~\cite{hinduja2024time} further argued that F1-binary is highly sensitive to class imbalance and threshold selection, advocating for more stable alternatives. These studies show that performance metrics can follow dataset-level distributional properties, but they don't analyze variability introduced by stochastic subject assignment within cross-validation.

\subsection{Cross-Domain Generalization and Bias}

Cross-dataset generalization remains a major challenge in AU detection. Ertugrul \emph{et al.}~\cite{crossing_domains_au_coding_pmc} analyzed cross-domain AU coding and demonstrated substantial degradation under distribution shift. Subsequent work has explored domain-incremental continual learning~\cite{domain_incremental_continual_learning_ieee}, identity-adversarial training~\cite{identity_adversarial_training_2025}, and dataset bias characterization~\cite{conceptscope_2025} to improve robustness. While these approaches address dataset-level distribution shift, they do not isolate performance variability from intra-dataset subject partitioning.
\section{Quantifying Split-Level Noise}
\label{sec:split_noise}

\subsection{Experimental Protocol}
BP4D+ because is a large, widely used AU benchmark, and many recent papers report within-dataset cross-validation on it; this makes it a representative baseline for quantifying protocol-induced noise. Considering this, all experiments are conducted on BP4D+ under a strictly subject-exclusive 3-fold cross-validation protocol.  More specifically, let $\mathcal{S}$ denote the set of subjects. For each trial, $\mathcal{S}$ is partitioned into three disjoint folds $\mathcal{S}_1, \mathcal{S}_2, \mathcal{S}_3$ such that $\mathcal{S}_i \cap \mathcal{S}_j = \emptyset$ for $i \neq j$. Training is performed on two folds and evaluation on the held-out fold, ensuring complete identity separation between train and test splits.
\begin{table}[t]
\centering
\caption{AU Prevalence Variation Across 3-Fold Subject Splits}
\label{tab:base_rate_variation}
\begin{tabular}{lccc}
\toprule
AU & Min & Max & Range \\
\midrule
1  & 0.0653 & 0.1217 & 0.0563 \\
2  & 0.0490 & 0.1016 & 0.0526 \\
4  & 0.0276 & 0.0870 & 0.0594 \\
6  & 0.4525 & 0.5398 & 0.0873 \\
7  & 0.6025 & 0.7212 & 0.1187 \\
10 & 0.6154 & 0.6859 & 0.0704 \\
12 & 0.5287 & 0.6340 & 0.1053 \\
14 & 0.5651 & 0.6446 & 0.0794 \\
15 & 0.0849 & 0.1214 & 0.0365 \\
17 & 0.0994 & 0.1614 & 0.0620 \\
23 & 0.1452 & 0.1856 & 0.0404 \\
24 & 0.0260 & 0.0550 & 0.0290 \\
\bottomrule
\end{tabular}
\end{table}

To isolate partition-induced stochasticity, four independent random subject partitions are generated. All models are trained with identical preprocessing, optimization, and a fixed decision threshold of 0.5. Metrics were computed independently per AU; no macro- or micro-averaging is used when estimating instability. To analyze split-induced variance independently of architecture, three backbone families were evaluated under identical conditions: ResNet50, MobileViT, and VGG16. 

\subsection{Distributional Perturbation from Subject Reassignment}

Subject-exclusive cross-validation preserves identity independence but does not preserve class distribution. Let $p_{au}^{(k)}$ denote the prevalence (base rate) of AU $au$ in fold $k$, defined as $p_{au}^{(k)} = \frac{\#\{y_{au}=1\}}{\#\{y_{au}\in\{0,1\}\}}$ computed over samples in that fold (excluding missing/invalid AU labels). The fold-induced distributional perturbation is $\Delta p_{au} = \max_k p_{au}^{(k)} - \min_k p_{au}^{(k)}$. From this, it can be seen (Table~\ref{tab:base_rate_variation}) that several AUs exhibit substantial prevalence shifts under 3-fold partitioning. For AU7 and AU12, the absolute prevalence range exceeds 0.10. Even low-prevalence AUs such as AU1 and AU4 show noticeable relative shifts. Consequently, each fold corresponds to a measurably different empirical test distribution. Because F1 is threshold-dependent and sensitive to base rates, this fold-level distributional variability propagates directly into performance variability. Thus, metric instability is structurally induced by subject reassignment, even under identical training conditions.

\subsection{Fold-Induced Performance Variance Across Backbones}
\label{sec:split_noise_variance}

Figure~\ref{fig:f1_range_all_backbones} visualizes the per-AU F1 distribution and Figure~\ref{fig:auc_range_all_backbones} does the same for AUC (Figure~\ref{fig:noise_ranges}a--b). Both are shown under 3-fold subject-exclusive evaluation for all three backbones. AUs are displayed on an equidistant categorical axis to emphasize variability magnitude rather than numeric AU indexing. For each AU and backbone, the marker denotes mean score, and vertical bars indicate the fold-to-fold min--max range. Several consistent patterns can be observed.

\begin{figure*}[!htp]
\centering

\begin{subfigure}[t]{\textwidth}
\caption{F1.}
\centering
\begin{tikzpicture}
\begin{axis}[
    width=\linewidth,
    height=5.6cm,
    xlabel={},
    ylabel={F1},
    ymin=0.1, ymax=1,
    symbolic x coords={1,2,4,6,7,10,12,14,15,17,23,24},
    xtick=data,
    xticklabels={}, 
    x tick label style={font=\small},
    y tick label style={font=\small},
    label style={font=\small},
    grid=major,
    grid style=dashed,
    legend style={
        at={(0.98,0.98)},
        anchor=north east,
        font=\small
    },
    enlarge x limits=0.10,
]

\addplot[
    only marks,
    mark=*,
    mark size=2.4pt,
    color=blue,
    xshift=-7pt,
    error bars/.cd,
        y dir=both,
        y explicit,
    error bar style={line width=1.0pt},
    error mark options={line width=1.0pt},
] coordinates {
    (1,0.44)  += (0,0.04) -= (0,0.03)
    (2,0.35)  += (0,0.04) -= (0,0.04)
    (4,0.32)  += (0,0.07) -= (0,0.06)
    (6,0.85)  += (0,0.02) -= (0,0.02)
    (7,0.88)  += (0,0.02) -= (0,0.02)
    (10,0.89) += (0,0.02) -= (0,0.02)
    (12,0.87) += (0,0.03) -= (0,0.03)
    (14,0.79) += (0,0.02) -= (0,0.02)
    (15,0.40) += (0,0.08) -= (0,0.07)
    (17,0.46) += (0,0.08) -= (0,0.05)
    (23,0.55) += (0,0.03) -= (0,0.03)
    (24,0.25) += (0,0.06) -= (0,0.07)
};
\addlegendentry{MobileViT (F1)}

\addplot[
    only marks,
    mark=*,
    mark size=2.4pt,
    color=orange,
    xshift=0pt,
    error bars/.cd,
        y dir=both,
        y explicit,
    error bar style={line width=1.0pt},
    error mark options={line width=1.0pt},
] coordinates {
    (1,0.45)  += (0,0.07) -= (0,0.11)
    (2,0.37)  += (0,0.05) -= (0,0.11)
    (4,0.33)  += (0,0.10) -= (0,0.07)
    (6,0.86)  += (0,0.02) -= (0,0.02)
    (7,0.89)  += (0,0.02) -= (0,0.02)
    (10,0.91) += (0,0.02) -= (0,0.02)
    (12,0.88) += (0,0.02) -= (0,0.03)
    (14,0.82) += (0,0.02) -= (0,0.03)
    (15,0.43) += (0,0.06) -= (0,0.09)
    (17,0.50) += (0,0.04) -= (0,0.08)
    (23,0.56) += (0,0.02) -= (0,0.03)
    (24,0.21) += (0,0.10) -= (0,0.11)
};
\addlegendentry{ResNet50 (F1)}

\addplot[
    only marks,
    mark=*,
    mark size=2.4pt,
    color=green!70!black,
    xshift=7pt,
    error bars/.cd,
        y dir=both,
        y explicit,
    error bar style={line width=1.0pt},
    error mark options={line width=1.0pt},
] coordinates {
    (1,0.44)  += (0,0.06) -= (0,0.09)
    (2,0.37)  += (0,0.05) -= (0,0.06)
    (4,0.33)  += (0,0.10) -= (0,0.11)
    (6,0.85)  += (0,0.02) -= (0,0.02)
    (7,0.90)  += (0,0.02) -= (0,0.02)
    (10,0.91) += (0,0.02) -= (0,0.02)
    (12,0.88) += (0,0.02) -= (0,0.03)
    (14,0.81) += (0,0.02) -= (0,0.03)
    (15,0.43) += (0,0.09) -= (0,0.08)
    (17,0.47) += (0,0.05) -= (0,0.06)
    (23,0.55) += (0,0.06) -= (0,0.05)
    (24,0.28) += (0,0.09) -= (0,0.16)
};
\addlegendentry{VGG16 (F1)}

\end{axis}
\end{tikzpicture}
\label{fig:f1_range_all_backbones}
\end{subfigure}


\begin{subfigure}[t]{\textwidth}
\centering
\begin{tikzpicture}
\begin{axis}[
    width=\linewidth,
    height=5.6cm,
    xlabel={AU},
    ylabel={AUC},
    ymin=0.10, ymax=1,
    symbolic x coords={1,2,4,6,7,10,12,14,15,17,23,24},
    xtick=data,
    x tick label style={font=\small},
    y tick label style={font=\small},
    label style={font=\small},
    grid=major,
    grid style=dashed,
    legend style={
        at={(0.98,0.02)},
        anchor=south east,
        font=\small
    },
    enlarge x limits=0.10,
]

\addplot[
    only marks,
    mark=*,
    mark size=2.4pt,
    color=blue,
    xshift=-7pt,
    error bars/.cd,
        y dir=both,
        y explicit,
    error bar style={line width=1.0pt},
    error mark options={line width=1.0pt},
] coordinates {
    (1,0.819)  += (0,0.015) -= (0,0.030)
    (2,0.776)  += (0,0.040) -= (0,0.029)
    (4,0.765)  += (0,0.034) -= (0,0.037)
    (6,0.925)  += (0,0.011) -= (0,0.016)
    (7,0.893)  += (0,0.016) -= (0,0.016)
    (10,0.926) += (0,0.020) -= (0,0.016)
    (12,0.931) += (0,0.016) -= (0,0.024)
    (14,0.756) += (0,0.030) -= (0,0.023)
    (15,0.804) += (0,0.032) -= (0,0.037)
    (17,0.822) += (0,0.008) -= (0,0.017)
    (23,0.841) += (0,0.020) -= (0,0.012)
    (24,0.843) += (0,0.048) -= (0,0.029)
};
\addlegendentry{MobileViT (AUC)}

\addplot[
    only marks,
    mark=*,
    mark size=2.4pt,
    color=orange,
    xshift=0pt,
    error bars/.cd,
        y dir=both,
        y explicit,
    error bar style={line width=1.0pt},
    error mark options={line width=1.0pt},
] coordinates {
    (1,0.831)  += (0,0.038) -= (0,0.028)
    (2,0.793)  += (0,0.030) -= (0,0.041)
    (4,0.812)  += (0,0.033) -= (0,0.039)
    (6,0.935)  += (0,0.008) -= (0,0.010)
    (7,0.909)  += (0,0.008) -= (0,0.009)
    (10,0.937) += (0,0.020) -= (0,0.016)
    (12,0.943) += (0,0.010) -= (0,0.011)
    (14,0.807) += (0,0.031) -= (0,0.022)
    (15,0.841) += (0,0.035) -= (0,0.031)
    (17,0.847) += (0,0.016) -= (0,0.016)
    (23,0.853) += (0,0.030) -= (0,0.022)
    (24,0.874) += (0,0.041) -= (0,0.059)
};
\addlegendentry{ResNet50 (AUC)}

\addplot[
    only marks,
    mark=*,
    mark size=2.4pt,
    color=green!70!black,
    xshift=7pt,
    error bars/.cd,
        y dir=both,
        y explicit,
    error bar style={line width=1.0pt},
    error mark options={line width=1.0pt},
] coordinates {
    (1,0.824)  += (0,0.041) -= (0,0.033)
    (2,0.785)  += (0,0.042) -= (0,0.038)
    (4,0.759)  += (0,0.083) -= (0,0.061)
    (6,0.930)  += (0,0.010) -= (0,0.011)
    (7,0.909)  += (0,0.012) -= (0,0.013)
    (10,0.937) += (0,0.015) -= (0,0.014)
    (12,0.941) += (0,0.009) -= (0,0.011)
    (14,0.782) += (0,0.019) -= (0,0.044)
    (15,0.826) += (0,0.018) -= (0,0.011)
    (17,0.839) += (0,0.016) -= (0,0.026)
    (23,0.843) += (0,0.029) -= (0,0.024)
    (24,0.871) += (0,0.053) -= (0,0.051)
};
\addlegendentry{VGG16 (AUC)}

\end{axis}
\end{tikzpicture}
\caption{AUC.}
\label{fig:auc_range_all_backbones}
\end{subfigure}
\caption{3-fold performance range per AU across backbones. Markers denote mean; vertical bars fold-to-fold min--max range.}
\label{fig:noise_ranges}
\end{figure*}
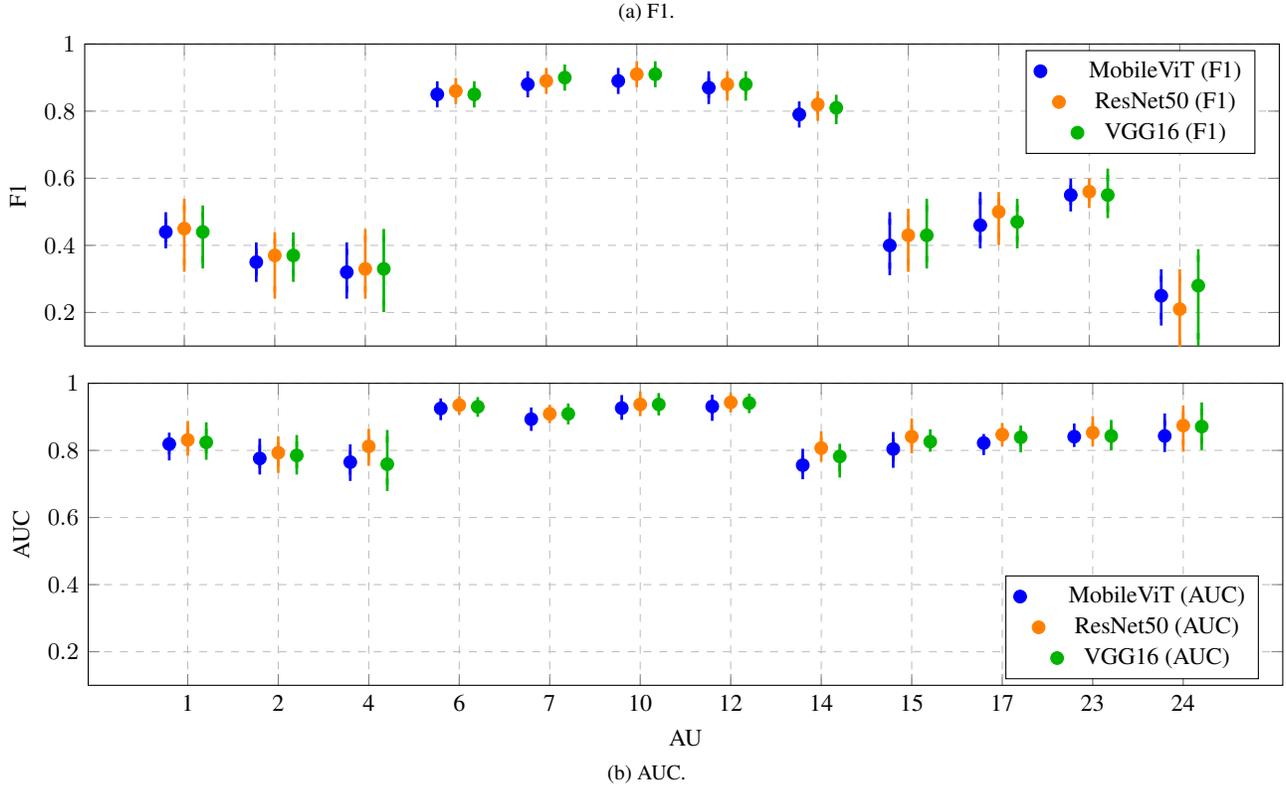
\begin{table*}[t]
\centering
\caption{Empirical noise floor per AU (ResNet50, 3-fold subject-exclusive): F1 and AUC.}
\label{tab:noise_floor}
\resizebox{\textwidth}{!}{%
\begin{tabular}{lcccccccc}
\toprule
& \multicolumn{4}{c}{F1} & \multicolumn{4}{c}{AUC} \\
\cmidrule(lr){2-5} \cmidrule(lr){6-9}
AU & Mean & $\sigma$ & 95\% Margin ($\pm1.96\sigma$) & Range & Mean & $\sigma$ & 95\% Margin ($\pm1.96\sigma$) & Range \\
\midrule
1  & 0.4540 & 0.0568 & $\pm0.1112$ & 0.3424--0.5280 & 0.8311 & 0.0193 & $\pm0.0379$ & 0.8006--0.8797 \\
2  & 0.3662 & 0.0476 & $\pm0.0933$ & 0.2623--0.4282 & 0.7926 & 0.0243 & $\pm0.0477$ & 0.7470--0.8252 \\
4  & 0.3294 & 0.0590 & $\pm0.1156$ & 0.2597--0.4278 & 0.8118 & 0.0297 & $\pm0.0583$ & 0.7705--0.8444 \\
6  & 0.8600 & 0.0108 & $\pm0.0212$ & 0.8418--0.8753 & 0.9355 & 0.0055 & $\pm0.0107$ & 0.9227--0.9443 \\
7  & 0.8862 & 0.0120 & $\pm0.0236$ & 0.8565--0.9057 & 0.9086 & 0.0054 & $\pm0.0106$ & 0.8982--0.9162 \\
10 & 0.9038 & 0.0102 & $\pm0.0200$ & 0.8922--0.9235 & 0.9370 & 0.0120 & $\pm0.0234$ & 0.9192--0.9571 \\
12 & 0.8937 & 0.0165 & $\pm0.0324$ & 0.8707--0.9214 & 0.9426 & 0.0087 & $\pm0.0171$ & 0.9311--0.9530 \\
14 & 0.8203 & 0.0188 & $\pm0.0368$ & 0.7919--0.8458 & 0.8072 & 0.0169 & $\pm0.0330$ & 0.7810--0.8384 \\
15 & 0.4357 & 0.0454 & $\pm0.0890$ & 0.3455--0.4968 & 0.8407 & 0.0227 & $\pm0.0445$ & 0.8064--0.8753 \\
17 & 0.5017 & 0.0285 & $\pm0.0558$ & 0.4632--0.5493 & 0.8469 & 0.0110 & $\pm0.0215$ & 0.8293--0.8632 \\
23 & 0.5617 & 0.0145 & $\pm0.0284$ & 0.5407--0.5857 & 0.8531 & 0.0189 & $\pm0.0370$ & 0.8301--0.8832 \\
24 & 0.2129 & 0.0797 & $\pm0.1562$ & 0.1042--0.3172 & 0.8737 & 0.0349 & $\pm0.0683$ & 0.8091--0.9185 \\
\bottomrule
\end{tabular}%
}
\end{table*}
First, instability is AU-dependent. Low-prevalence AUs (e.g., AU24, AU4, AU1, AU15) exhibit substantially larger vertical ranges than high-prevalence AUs (e.g., AU6, AU7, AU10, AU12). For AU24, the fold-induced F1 span exceeds 0.15 for every backbone; for ResNet50, the range is [0.1042, 0.3172] (Table~\ref{tab:noise_floor}), matching the tallest bars in Figure~\ref{fig:f1_range_all_backbones}. AU4 and AU15 show similarly wide spreads. In contrast, AU6 and AU10 remain comparatively stable. Second, instability is consistent across architectures. Although mean performance differs slightly between ResNet50, MobileViT, and VGG16, the  \textit{relative volatility pattern} is preserved: the same AUs that are unstable under one backbone are unstable under the others. This indicates that the dominant source of variability is the data partition rather than the network architecture. Third, backbone capacity does not eliminate split sensitivity. ResNet50 does not uniformly reduce fold variance relative to VGG16 or MobileViT; in some AUs, the range is comparable or even larger. This suggests that increasing representational capacity does not inherently stabilize evaluation under subject reassignment.

Formally, let $m_{au}^{(k)}$ denote the F1 score for AU under fold $k$. Split-induced variability is quantified using the standard deviation across all folds, which allows us to establish a 95\% empirical noise margin (Section 3.4). Figure~\ref{fig:f1_range_all_backbones} makes this noise floor visually explicit: for several AUs, the fold-to-fold F1 variation exceeds the magnitude of typical incremental improvements reported as state-of-the-art. The visual spans in Figure~\ref{fig:f1_range_all_backbones} correspond to the per-AU min--max ranges reported in Table~\ref{tab:noise_floor}. Importantly, this variance arises without altering model architecture, hyperparameters, or dataset composition; only subject allocation changes.

\subsection{Defining the Noise Floor}
To quantify evaluation uncertainty, we treat cross-validation performance as a random variable conditioned on subject partitioning. For each AU, we compute the standard deviation of F1 across all 3-fold splits, $\sigma_{au} = \mathrm{StdDev}(m_{au}^{(1)}, \ldots, m_{au}^{(K)})$, and define the per-AU 95\% \emph{instability margin} as $\text{Noise}_{au} = \pm 1.96\,\sigma_{au}$, which corresponds to the 95\% interval under repeated random subject allocation. We use \emph{noise floor} to denote the \textbf{protocol-level average} of these per-AU 95\% margins (mean of $\pm1.96\sigma_{au}$ across AUs); we refer to this as the \emph{average 95\% margin}.

Table~\ref{tab:noise_floor} reports the statistics for ResNet50 and it can be seen that the uncertainty varies dramatically across AUs. For AU24, the 95\% noise margin is ±0.156.  
For AU4 and AU1, it exceeds ±0.11.  
In contrast, AU6 and AU10 remain near ±0.02. The average 95\% margin across AUs (the protocol-level noise floor) is $\pm 0.065$ for ResNet50 under 3-fold subject-exclusive evaluation. This suggests that for several AUs, the evaluation protocol alone introduces uncertainty exceeding 0.10 absolute F1. Any reported improvement smaller than this margin cannot be separated from partition-induced variability under the same protocol.

\subsection{Backbone-Level Stability Comparison}

To determine whether split-level instability is architecture-specific or structural, we compute the standard deviation of F1 across all completed 3-fold splits for each backbone. For backbone $b$ and AU $i$, we define: $\sigma_{b,i} = \mathrm{StdDev}(F1_{b,i}^{(1)}, \dots, F1_{b,i}^{(K)})$,
where $K$ denotes all fold instances across completed splits.

Moving from 3-fold to 5-fold CV increases instability for both metrics under ResNet50 on BP4D+. Specifically, the mean per-AU standard deviation rises from 0.0333 to 0.0506 for F1 (corresponding to a 95\% margin of $\pm 0.065$ vs. $\pm 0.099$), and from 0.0174 to 0.0261 for AUC (95\% margin $\pm 0.034$ vs. $\pm 0.051$). The increase in noise is expected: for a fixed underlying distribution, the variance of estimated metrics scales inversely with test-set size ($\mathrm{Var} \propto 1/n$). Moving from 33\% test coverage (3-fold) to 20\% (5-fold) reduces $n$ per fold, which amplifies per-fold estimation variance and, consequently, the observed cross-fold $\sigma$.

All three architectures exhibit substantial split-induced variance. The mean per-AU standard deviation ranges from 0.0308 to 0.0368 across backbones. Even the most stable architecture (MobileViT) shows non-trivial volatility. VGG16 exhibits the highest overall instability, both in mean and maximum $\sigma$. ResNet50 shows the single most extreme AU-level instability (AU24), while MobileViT remains slightly more stable but still clearly affected. These results suggest that split-level noise is not an artifact of a particular model design. Rather, it emerges from subject reassignment itself. Architectural differences modulate the magnitude of instability but do not eliminate it (see Section~\ref{sec:split_noise_variance}).

\subsection{Metric-Level Instability: F1 vs. AUC}
\label{sec:metric_instability}

We next examine whether split-induced instability depends on the evaluation metric. All previous analysis used F1 at a fixed threshold. Because F1 is threshold-dependent and directly influenced by the number of positive examples in the test fold, it may amplify the distributional perturbations introduced by subject reassignment. In contrast, AUC evaluates ranking quality across all thresholds and is less sensitive to prevalence shifts. To quantify this effect, we compute the standard deviation of both F1 and AUC across all completed 3-fold splits for each AU (Table~\ref{tab:f1_auc_volatility}).
\begin{table}[t]
\centering
\caption{Per-AU volatility comparison between F1 and AUC for ResNet50 under 3-fold subject-exclusive cross-validation. Volatility is measured as the cross-split standard deviation $\sigma$ of each metric; the volatility ratio $\rho = \sigma_{F1}/\sigma_{AUC}$ indicates how much more (or less) F1 fluctuates than AUC.}
\label{tab:f1_auc_volatility}
\begin{tabular}{lccc}
\toprule
AU & $\sigma_{F1}$ & $\sigma_{AUC}$ & Volatility Ratio $\rho$ \\
\midrule
1  & 0.0568 & 0.0193 & 2.93 \\
2  & 0.0476 & 0.0243 & 1.96 \\
4  & 0.0590 & 0.0297 & 1.99 \\
6  & 0.0108 & 0.0055 & 1.98 \\
7  & 0.0120 & 0.0054 & 2.24 \\
10 & 0.0102 & 0.0120 & 0.85 \\
12 & 0.0165 & 0.0087 & 1.89 \\
14 & 0.0188 & 0.0169 & 1.11 \\
15 & 0.0454 & 0.0227 & 2.00 \\
17 & 0.0285 & 0.0110 & 2.59 \\
23 & 0.0145 & 0.0189 & 0.77 \\
24 & 0.0797 & 0.0349 & 2.29 \\
\bottomrule
\end{tabular}

\end{table}

For most AUs, F1 is substantially more volatile than AUC. The volatility ratio exceeds 2.0 for AU1, AU4, AU7, AU15, AU17, and AU24. For AU1, F1 fluctuates nearly three times as much as AUC. Only two AUs (AU10 and AU23) exhibit slightly higher AUC variance. This pattern reflects how the metrics respond to fold composition. When subject reassignment changes AU prevalence in the test fold, F1 shifts accordingly because it operates at a fixed decision threshold. AUC, by integrating over all thresholds, dampens this sensitivity unless ranking structure itself changes. The narrower vertical bars in Figure~\ref{fig:auc_range_all_backbones} compared to Figure~\ref{fig:f1_range_all_backbones} visually mirror the volatility ratios in Table~\ref{tab:f1_auc_volatility}.

The implication is practical. If improvements are reported in terms of small F1 gains (e.g., +0.01 or +0.02), those gains may fall entirely within split-induced F1 volatility. Metric choice alone can therefore determine whether an apparent improvement appears statistically meaningful. Split-level noise is not only a property of the dataset or the model, it is also amplified by threshold-based evaluation.

\subsection{Statistical Implication for SOTA Claims}
\label{sec:sota_implication}

The empirical noise floor measured in Section~\ref{sec:split_noise} provides a useful reference point for interpreting reported improvements. Under repeated 3-fold subject-exclusive splits, the average 95\% margin (noise floor) is approximately $\pm 0.065$ in F1, computed as the mean of the per-AU margins in Table~\ref{tab:noise_floor}. For several AUs, the margin is substantially larger. This means that performance differences smaller than  $\approx0.06$ average F1 may not be reliably distinguishable from split-induced variation under the same protocol.

Representative AU detection results on BP4D+ from earlier CNN-based methods through recent transformer and masked pretraining approaches are summarized in Table~\ref{table:bp4d_plus_full}. All methods fall within the empirical split-level noise band of $\pm 0.065$ average F1, suggesting that fine-grained ranking among strong models can be unstable under subject partitioning. The spread from the best to the worst entry is 0.041 F1 (0.668 to 0.627), and the gap between the top result and the median is 0.019 F1 (0.668 vs. 0.649). Both are smaller than the protocol-level noise band, indicating that much of the apparent ranking within this range is not reliably distinguishable under repeated splits.

\begin{table}[t]
\centering
\caption{State-of-the-art AU detection performance on BP4D+. Entries list the reported mean F1 (average over AUs) for representative methods by year under within-dataset subject-exclusive evaluation; the full spread of results falls within the empirical split-level noise band of $\pm 0.065$ average F1.}
\label{table:bp4d_plus_full}
\begin{tabular}{lcc}
\hline
\textbf{Model} & \textbf{Year} & \textbf{BP4D+ (F1)} \\ \hline
VisAULa \cite{visaula_2026} & 2026 & 0.667 \\
STIF-DA \cite{stifda2025} & 2025 & 0.654 \\
FMAE-IAT \cite{fmae_2024} & 2024 & 0.668 \\
MDHR \cite{Wang2024MDHR} & 2024 & 0.666 \\
AUFormer \cite{auformer_2024} & 2024 & 0.662 \\
MCM \cite{mcm2024} & 2024 & 0.66 \\
AQ-CSL \cite{aqcsl2024} & 2024 & 0.649 \\
MS-PSAC \cite{mspsac2024} & 2024 & 0.646 \\
FG-Net \cite{fgnet2024} & 2024 & 0.638 \\
CLEF \cite{clef2023} & 2023 & 0.659 \\
OpenGraph \cite{opengraph2023} & 2023 & 0.640 \\
ANFL \cite{anfl2022} & 2022 & 0.655 \\
KSRL \cite{ksrl2022} & 2022 & 0.645 \\
ME-Graph \cite{megraph2022} & 2022 & 0.633 \\
PIAP \cite{piap2021} & 2021 & 0.644 \\
FAUT \cite{faut2021} & 2021 & 0.642 \\
JAA-Net \cite{jaa_net_eccv2018} & 2021 & 0.627 \\
\hline
\end{tabular}
\end{table}

The long-term trend indicates substantial improvement from early CNN baselines in the mid-50s F1 range to modern transformer and masked-pretraining approaches in the mid-60s. That jump is large and well beyond the empirical noise margin, indicating genuine progress over time. However, once performance reaches the low-to-mid 60s, improvements become much smaller. Most successive gains are between +0.005 and +0.026 F1, all of which fall below our established average noise band. This does not suggest that architectural refinements are ineffective. Instead, it highlights a limitation of the evaluation protocol: when improvements fall below the noise margin, model ordering may depend on fold composition (see Section~\ref{sec:split_noise_variance}). As margins shrink and benchmarks saturate, evaluation rigor must increase accordingly. Reporting variance across repeated splits and complementing intra-dataset cross-validation with cross-dataset protocols such as Leave-One-Dataset-Out (LODO) become necessary to reliably distinguish small performance differences.

Beyond split-level instability, another structural limitation of current AU evaluation practice is the reliance on single-dataset cross-validation. Most recent SOTA methods are evaluated exclusively on BP4D+ using subject-exclusive folds, without reporting out-of-dataset generalization performance. As a result, improvements are measured within a fixed distribution and under a protocol that we have shown to exhibit non-negligible stochastic variance. This raises a broader question: even if two models are statistically indistinguishable under repeated splits, do they generalize differently to unseen domains? To address both partition-level instability and domain-level robustness, we next consider Leave-One-Dataset-Out (LODO) evaluation.

\section{Leave-One-Dataset-Out Evaluation}
\label{sec:lodo}
\begin{table}[t]
\centering
\caption{AU prevalence across datasets. ``--'' indicates AU not annotated in the corresponding dataset.}
\label{tab:lodo_prevalence}
\resizebox{0.98\linewidth}{!}{%
\begin{tabular}{lccccc}
\toprule
AU & BP4D & BP4D+ & DISFA & GFT & UNBC \\
\midrule
1  & 0.211 & 0.097 & 0.067 & 0.075 & -- \\
2  & 0.171 & 0.082 & 0.056 & 0.105 & -- \\
4  & 0.203 & 0.058 & 0.188 & 0.041 & 0.022 \\
6  & 0.461 & 0.498 & 0.149 & 0.288 & 0.115 \\
7  & 0.549 & 0.663 & --    & 0.369 & 0.070 \\
10 & 0.594 & 0.648 & --    & 0.321 & 0.011 \\
12 & 0.562 & 0.579 & 0.235 & 0.264 & 0.142 \\
14 & 0.466 & 0.601 & --    & 0.526 & -- \\
15 & 0.169 & 0.107 & 0.060 & 0.079 & -- \\
17 & 0.343 & 0.130 & 0.099 & 0.254 & -- \\
23 & 0.165 & 0.167 & --    & 0.234 & -- \\
24 & 0.151 & 0.039 & --    & 0.116 & -- \\
\bottomrule
\end{tabular}}
\end{table}
Section~\ref{sec:split_noise} showed that subject-exclusive cross-validation within a single dataset introduces measurable stochastic variance. We now move to a stricter protocol that removes partition randomness entirely and directly evaluates cross-dataset generalization: Leave-One-Dataset-Out (LODO).

\subsection{Datasets and Base-Rate Differences}

We consider five widely used AU datasets: BP4D \citep{bp4d_zhang_2014}, BP4D+ \citep{zhang2016multimodal}, DISFA \citep{disfa_mavadati_2013}, Sayette GFT \citep{gft_girard_2017}, and UNBC Pain Archive \citep{unbc_lucey_2011}. These datasets differ in elicitation protocol (posed vs.\ spontaneous vs.\ pain-related), recording conditions, subject populations, and annotation density. Importantly, they also differ substantially in AU prevalence. Table~\ref{tab:lodo_prevalence} reports the per-dataset base rates for the 12 commonly evaluated AUs, and the differences are non-trivial. For example, AU6 and AU7 are highly prevalent in BP4D+ but considerably less frequent in DISFA and UNBC. AU24 is relatively rare in BP4D+ and absent in other datasets. These shifts imply that each dataset defines a distinct statistical regime, so cross-dataset performance should be interpreted with base-rate differences in mind. It is also important to note that some AUs are not annotated in all datasets.

\begin{table*}[t]
\centering
\caption{\textbf{LODO performance of the enhanced model.} Each Test dataset is held out once. Performance is grouped by metric, with the Metric column indicating whether the row reports F1-score or AUC. Results are based on a fixed classification threshold of 0.5.}
\label{tab:lodo_all}
\resizebox{\textwidth}{!}{
\begin{tabular}{clcccccccccccc}
\toprule
Metric & Test Dataset & AU1 & AU2 & AU4 & AU6 & AU7 & AU10 & AU12 & AU14 & AU15 & AU17 & AU23 & AU24 \\
\midrule
\multirow{5}{*}{\textbf{F1}} 
 & BP4D    & 0.537 & 0.496 & 0.549 & 0.797 & 0.803 & 0.847 & 0.863 & 0.686 & 0.484 & 0.546 & 0.507 & 0.238 \\
 & BP4D+   & 0.368 & 0.334 & 0.343 & 0.848 & 0.852 & 0.885 & 0.876 & 0.669 & 0.374 & 0.438 & 0.509 & 0.328 \\
 & DISFA   & 0.391 & 0.296 & 0.514 & 0.588 & --    & --    & 0.760 & --    & 0.370 & 0.325 & --    & --    \\
 & GFT     & 0.221 & 0.186 & 0.112 & 0.742 & 0.556 & 0.527 & 0.769 & 0.084 & 0.168 & 0.210 & 0.179 & 0.146 \\
 & UNBC    & --    & --    & 0.095 & 0.419 & 0.224 & 0.036 & 0.485 & --    & 0.001 & --    & --    & --    \\
\midrule
\multirow{5}{*}{\textbf{AUC}} 
 & BP4D    & 0.817 & 0.775 & 0.827 & 0.893 & 0.839 & 0.869 & 0.937 & 0.717 & 0.828 & 0.779 & 0.803 & 0.831 \\
 & BP4D+   & 0.799 & 0.803 & 0.766 & 0.931 & 0.882 & 0.907 & 0.927 & 0.699 & 0.788 & 0.812 & 0.810 & 0.879 \\
 & DISFA   & 0.772 & 0.781 & 0.758 & 0.868 & --    & --    & 0.938 & --    & 0.725 & 0.715 & --    & --    \\
 & GFT     & 0.725 & 0.652 & 0.741 & 0.909 & 0.790 & 0.777 & 0.922 & 0.547 & 0.668 & 0.593 & 0.568 & 0.664 \\
 & UNBC    & --    & --    & 0.774 & 0.837 & 0.798 & 0.776 & 0.833 & --    & 0.827 & --    & --    & --    \\
\bottomrule
\end{tabular}
}
\end{table*}

\begin{figure}
    \centering
    \includegraphics[width=\linewidth]{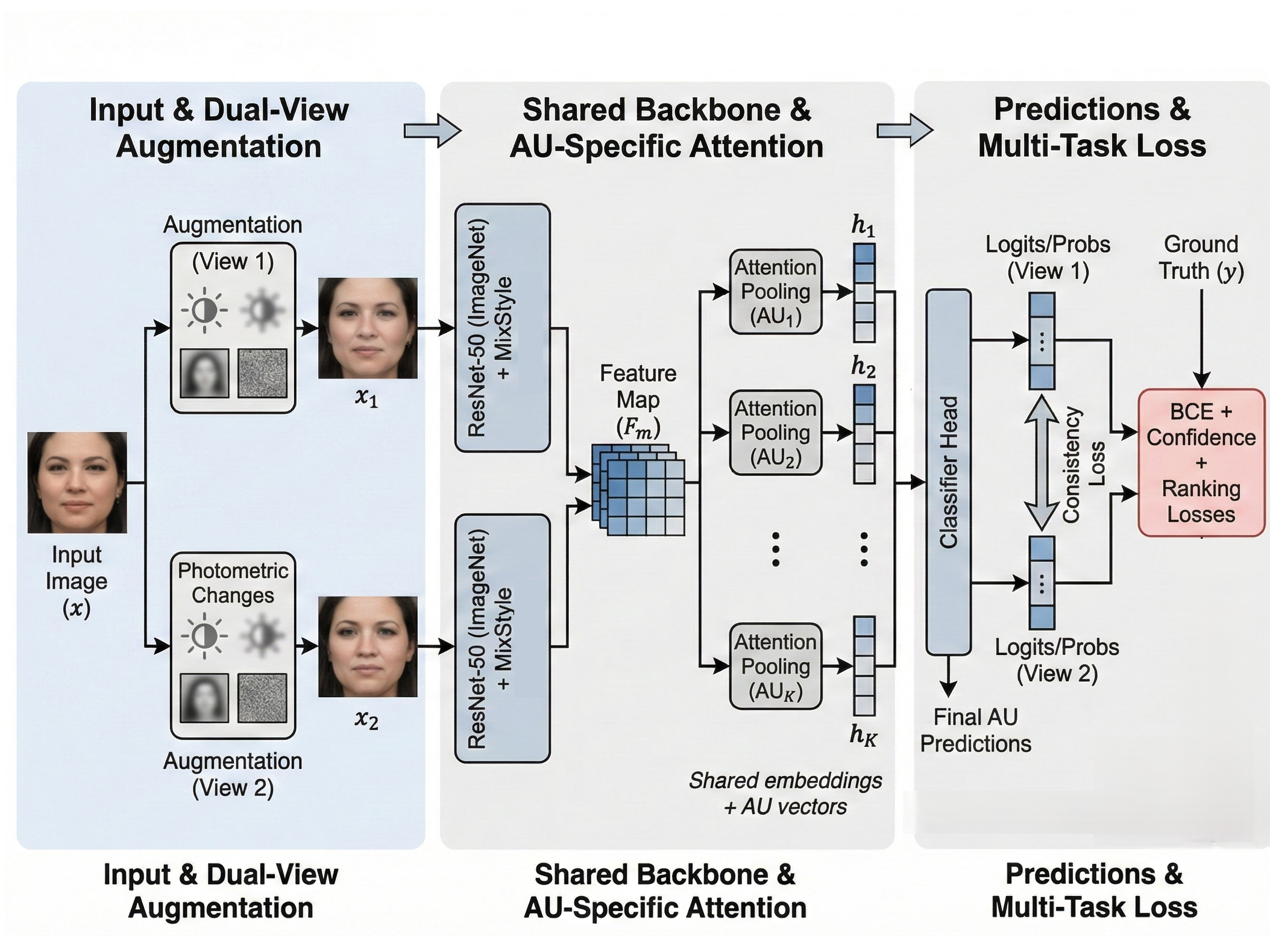}
    \caption{Overview of the LODO training pipeline. }
    \label{fig:lodo_pipeline}
\end{figure}
\subsection{LODO Protocol}
The proposed end-to-end LODO workflow is shown in Figure~\ref{fig:lodo_pipeline}. The backbone produces shared feature maps, which feed AU-specific attention pooling modules that focus on localized regions before classification. MixStyle blocks inside the backbone perturb feature statistics to simulate domain shifts during training. The right panel highlights that the classifier is evaluated only on the held-out dataset, with no target-domain statistics used for calibration. This visualization emphasizes the train-on-pooled, test-on-held-out structure of the LODO protocol and the points where domain robustness is encouraged. Let $\mathcal{D} = \{D_1, \dots, D_K\}$ denote the set of datasets. Under LODO, one dataset $D_k$ is held out entirely for testing, while the model is trained on the pooled union of the remaining datasets: $\text{Train} = \bigcup_{i \neq k} D_i$, $\text{Test} = D_k$. No subjects, frames, statistics, or validation signals from the held-out dataset are used during training or hyperparameter selection. This eliminates fold-induced stochasticity and prevents dataset-specific threshold calibration. The process is repeated for each dataset.

\subsection{Model and Training Setup}
A shared ResNet50 backbone (Figure~\ref{fig:lodo_pipeline}), initialized with ImageNet weights, extracts spatial feature maps from each input image. 
For each AU, a dedicated attention pooling module learns to focus on AU-relevant regions of the shared feature representation. 
The pooled AU-specific embeddings are passed through a multi-label classifier head to produce per-AU logits. We incorporate MixStyle layers within intermediate residual blocks to encourage robustness to domain shifts by perturbing feature statistics during training. This is performed on the pooled union of all non-held-out datasets using the multi-task objective, which combines binary cross-entropy, confidence regularization, ranking losses, and view-consistency constraints. A fixed decision threshold of 0.5 is used at evaluation time.

\subsection{Evaluation Metrics}
We report both AUC and F1 per AU, which allows us to separate threshold-independent representation stability (AUC) from threshold-sensitive operating-point performance (F1) under cross-dataset transfer.
\section{Results}
\label{sec:lodo_results}

\subsection{LODO Performance}
\label{subsec:lodo_enhanced}

Table~\ref{tab:lodo_all} presents the full Leave-One-Dataset-Out results for the enhanced model. 
Each dataset is excluded once during training and used only for testing. 
A fixed threshold of 0.5 is applied across all datasets without recalibration. 
Both F1 and AUC are reported for all annotated AUs.

The difficulty of cross-dataset generalization is evident in the magnitude of variation across held-out datasets. 
When BP4D or BP4D+ is excluded during training, most AUs retain moderate-to-strong performance. 
When UNBC is held out, several AUs exhibit near-zero F1, even though AUC remains defined for some of them. 
This reflects extreme prevalence sparsity in UNBC rather than a complete loss of discriminative signal.

AUC remains noticeably more stable than F1 across transfers. 
For example, under the UNBC condition, AU6 achieves an AUC of 0.837 while its F1 is 0.419. 
AU12 maintains AUC values above 0.92 across multiple held-out datasets, while its F1 fluctuates more substantially. 
This behavior is consistent with the split-level analysis earlier in the paper: ranking ability degrades more slowly than threshold-dependent performance under distribution shift.

High-prevalence AUs such as AU6, AU7, AU10, and AU12 remain comparatively stable across datasets. Lower-prevalence AUs such as AU15, AU17, and AU24 show much larger variability. For AU24, F1 differs sharply across held-out datasets, while AUC remains measurable whenever annotations exist. This indicates that operating-point instability is strongly coupled to base-rate differences. The variability observed here is fundamentally different from the split-induced noise (Section~\ref{sec:split_noise}). In that case, instability arose from subject reassignment within a fixed dataset. Here, the shifts reflect real differences in data distributions across domains. LODO therefore reveals cross-dataset sensitivity that standard cross-validation cannot capture. Taken together, the results show that even with domain-aware training, cross-dataset robustness remains limited. Ranking performance is comparatively stable, but operating-point performance varies substantially across domains.

\subsection{Subject-Level Bootstrap Analysis under LODO}
\label{subsec:lodo_bootstrap}

To quantify the stability of cross-dataset transfer, we perform subject-level bootstrapping for each LODO evaluation. By resampling subjects with replacement (1,000 bootstrap iterations per transfer) and recomputing metric differences across domains, we isolate whether observed performance shifts reflect systematic domain effects or merely sampling variability among subjects. We characterize this stability via \textit{Domain Sensitivity} (DS), defined as the percentage of cross-dataset transfers in which the performance shift is statistically significant ($p < 0.05$) under the bootstrap distribution. Significance is assessed with a two-sided percentile test: a transfer is counted as significant if 0 lies outside the 95\% bootstrap confidence interval of $\Delta$. There are five LODO evaluations (one per held-out dataset). DS is computed over the subset of transfers where the AU is annotated in the target dataset. Table~\ref{tab:lodo_bootstrap_summary} summarizes these aggregate statistics across all dataset pairings.

\begin{table}[t]
\centering
\caption{\textbf{Domain Sensitivity (DS) across LODO transfers.} Mean $\Delta$ is the \emph{signed} average shift in metric when moving from pooled-source evaluation to the held-out target dataset (target minus source mean). DS measures the frequency of statistically significant ($p < 0.05$) performance changes under subject-level bootstrapping. F1 exhibits significantly higher sensitivity to domain shifts than AUC.}
\label{tab:lodo_bootstrap_summary}
\resizebox{0.98\linewidth}{!}{
\begin{tabular}{lcccc}
\toprule
AU & Mean $\Delta$F1 & Mean $\Delta$AUC & Mean F1 DS & Mean AUC DS \\
\midrule
1  & -0.172 & -0.065 & 67\% & 33\% \\
2  & -0.134 & +0.003 & 33\% & 0\%  \\
4  & -0.198 & -0.028 & 67\% & 0\%  \\
6  & -0.172 & -0.028 & 100\% & 60\% \\
7  & -0.386 & -0.027 & 100\% & 67\% \\
10 & -0.472 & -0.071 & 100\% & 50\% \\
12 & -0.167 & -0.044 & 80\% & 40\% \\
14 & -0.307 & -0.119 & 67\% & 67\% \\
15 & -0.242 & -0.011 & 50\% & 0\%  \\
17 & -0.147 & -0.042 & 100\% & 0\%  \\
23 & +0.002 & +0.007 & 0\%  & 0\%  \\
24 & +0.090 & +0.048 & 0\%  & 0\%  \\
\bottomrule
\end{tabular}}
\end{table}

The results reveal a contrast between ranking and classification stability. Across nearly all AUs, the magnitude of cross-dataset change in F1 significantly exceeds that of AUC. Notably, several AUs (AUs 6, 7, 10, and 17) exhibit an F1 Domain Sensitivity of 100\%, meaning the classification performance was significantly disrupted in all transfer pairings. In contrast, AUC shifts are less frequent and typically smaller in magnitude. This discrepancy confirms that operating-point performance (F1) is substantially more fragile than ranking performance (AUC) when encountering new domains. Even in cases where AUC remains relatively stable, F1 can fluctuate sharply, suggesting that the optimal decision threshold is highly domain-dependent. This domain-induced instability is distinct from the stochastic noise observed in within-dataset splits (Section ~\ref{sec:split_noise}), reinforcing the need for caution when interpreting operating-point improvements under cross-dataset evaluation.
\section{Implications for Evaluation Protocols}
\label{sec:implications}

Sections~\ref{sec:split_noise} and~\ref{sec:lodo_results} quantify two distinct sources of variability: split-level stochasticity within a dataset and domain shift across datasets. Together, they show that single-dataset cross-validation is insufficient for stable model ordering and that evaluation protocol choice can dominate small performance deltas. Considering this, we proposed the following practical recommendations for AU evaluation.
\begin{itemize}
    \item Report variance across repeated splits and include confidence intervals or statistical tests.
    \item Complement intra-dataset cross-validation with multi-dataset protocols (e.g., LODO) to measure robustness to domain shift.
    \item Report both threshold-free and operating-point metrics, and document any thresholding or calibration procedure.
    \item Treat gains that fall within the empirical noise band as statistically indistinguishable rather than definitive SOTA advances.
\end{itemize}
These recommendations shift emphasis from single-run rankings to reproducible, protocol-aware comparisons.
\section{Limitations and Future Work}
\label{sec:limitations}

This work empirically showed how subject-exclusive cross validation within a single dataset introduces measurable stochastic variance. To remove partition-level randomness, we propose LODO Evaluation.  Initial results using this approach are encouraging. 

Nevertheless, several limitations should be acknowledged.    First, although LODO removes partition-level randomness, it still depends on the specific set of datasets included. 
The five datasets used here (BP4D, BP4D+, DISFA, GFT, UNBC) represent a range of recording conditions and annotation styles, but they do not exhaust the diversity of real-world settings. 
As additional AU datasets become available, LODO benchmarks should be expanded. Second, our LODO experiments currently report results for a single model. While this is sufficient to establish a protocol baseline, a broader architectural comparison under LODO would strengthen conclusions about model design and domain invariance. Finally, the present analysis separates split-level noise and cross-dataset instability, but it does not yet explore their interaction under larger cross-validation schemes (e.g., repeated multi-fold CV across multiple datasets). 
Extending the noise-floor analysis to larger fold counts and additional training configurations would further clarify how evaluation variance scales with protocol design.

To address these limitations, future work would (1) complete all LODO folds; (2) compare against single-dataset and pairwise cross-dataset baselines; and (3) conduct ablations on backbone choice, domain randomization mechanisms, and training dataset composition. In this way, we can better determine whether improvements in AU detection reflect genuine progress or only evaluation artifacts.
{
    \small
    \bibliographystyle{ieeenat_fullname}
    \bibliography{main}
}


\end{document}